# A Conjoint Graph Representation Learning Framework for Hypertension Comorbidity Risk Prediction

*Leming Zhou[1]\*, Zuo Wang[2], Zhixuan Duan[2]*

[1]*College of Computer Science and Technology,*
*Chongqing University of Posts and Telecommunications, Chongqing, China*
[2]*College of Computer and Information Science, Southwest University, Chongqing, China*
*\* ywkzlm@126.com*

**Abstract**

The comorbidities of hypertension impose a heavy burden on patients and society. Early identification is necessary to prompt intervention, but it remains a challenging task. This study aims to address this challenge by combining joint graph learning with network analysis. Motivated by this discovery, we develop a Conjoint Graph Representation Learning (CGRL) framework that: a) constructs two networks based on disease coding, including the patient network and the disease difference network. Three comorbidity network features were generated based on the basic difference network to capture the potential relationship between comorbidities and risk diseases; b) incorporates computational structure intervention and learning feature representation, CGRL was developed to predict the risks of diabetes and coronary heart disease in patients; and c) analysis the comorbidity patterns and exploring the pathways of disease progression, the pathological pathogenesis of diabetes and coronary heart disease may be revealed. The results show that the network features extracted based on the difference network are important, and the framework we proposed provides more accurate predictions than other strong models in terms of accuracy.

## 1 Introduction

Hypertension and its associated cardiovascular metabolic comorbidities (including complications) are prevalent and pose significant risks. The annual number of deaths from cardiovascular disease is reported to increase from 12.4 million to 19.8 million between 1990 and 2022 [1]. According to the China Cardiovascular Health and Disease Report 2022, there are 330 million patients with cardiovascular disease in China, including 11.39 million with coronary heart disease (CHD), 8.9 million with heart failure, and 245 million with hypertension. The cardiovascular complications of diabetes mellitus (DM) patients have become a serious burden for patients and their families [2]. Hypertension and its series of complications are the main causes of death worldwide, such as the clustering of hypertension, type 2 diabetes, and dyslipidaemia, a heavy burden. In most elderly patients, hypertension, type 2 diabetes, coronary artery disease, cancer [3], and other related risk factors accelerate cardiovascular aging and lead to heart failure. From the perspective of avoiding harm, accurately predicting the risk of a certain disease outcome in advance is particularly important.

In recent years, technologies that are based on Artificial Intelligence (AI) have emerged to be powerful in predicting the risk of disease outcomes from the perspective of historical data, especially historical medical diagnostic data [3], [4]. The existing methods are usually based on the obtained disease vector features of patients as the input feature, and do not consider the extraction of features from the perspective of the difference between the individual network and the whole network, or only consider the test features for prediction and ignore the prediction based on the comorbidity network.

The first difficulty is in obtaining the medical data. The limited size of medical data is a common challenge in medical data analysis research [5]. Pre-processing work, such as data quality, cleaning, inclusion, and exclusion, can reduce the available amount of data. Second, to directly classify and predict based on high-dimensional disease features without extracting the differences between individual comorbidity networks and high-risk comorbidity networks makes it difficult to provide convincing explanations [6], [7].

To cope with the limitation mentioned, in this paper, we propose a novel graph-learning-based framework, namely the Conjoint Graph Representation Learning (CGRL) framework, which can learn features from the differential comorbidity network.

The main contributions of this paper are as follows:
- We construct the patient network based on common diseases from cohorts of two groups of retrospective patients, where the nodes and edges represent the patients and the number of common diseases between different patients, respectively.
- We present a network feature extraction method based on the differential comorbidity network. This constructed network is interpretable for comorbidity progression, thus rationalizing the prediction and being used for the subsequent risk prediction.

- We propose a risk prediction learning method of CGRL for hypertension comorbidities and perform experiments on the DM and CHD datasets, which demonstrate that CGRL can significantly enhance the accuracy of risk prediction.

## 2. Related work

To discover the comorbidity patterns and predict the risk disease, many methods for comorbidity network analysis and prediction of risk disease have been proposed.

A large number of studies [4], [6] have been conducted to study the pattern of comorbidities, linking diseases to each other to construct comorbidity networks that exhibit high comorbidity patterns, revealing hidden connections between different diseases. Researchers have also studied comorbidity patterns based on the Phenotypic Disease Network (PDN), a paired disease network constructed by using seven co-occurrence indicators in [8]. Due to the possibility of creating a co-occurrence network with multiple interactions, community detection methods have been combined to identify different communities of diseases that co-occur with colon cancer [9]. A latent factor (LF)-based approach is used to deal with the problem of incomplete data to predict missing data, which is significant for the missing values of medical test features [10]-[30]. Scholars have proposed non-negative matrix factorization methods [31]-[55] for community detection. Despite these studies, features extracted from comorbidity networks were not considered as features of individual patients.

Many studies have explored the relationship between patient comorbidities and predicted classification by calculating the similarity between patients. To learn useful node representations from the network, the graph neural network method has also been used to classify patients. Compared with machine learning [56]-[60], graph neural networks [61]-[71] have shown more advantages in integrating node features and network topology in building a comorbidity-based patient similarity network. However, the traditional GAT is only based on feature attention and neglects the structure attention, resulting in limited prediction ability. Our CGRL method overcomes these limitations by introducing a network feature extraction method based on a differential comorbidity network, and incorporating the Conjoint Graph Representation Learning (CGRL) framework to predict the hypertension comorbidity risk.

## 3. Background

In this section, we introduced the formulation of the preparatory knowledge and symbols for the DM and CHD prediction problem.

In this paper, we assume the patient weight network G={$V$, $E$}, where $V$ and $E$ represent patient node and edge set, including $Z$ patient nodes, $|E|$ edges represent the number of common diseases between different patients, and $CL$ represents the number of categories in the classification (2 classes, $CL$ is less than $Z$). $\mathbf{X} \in \mathbb{R}^{Z \times D}$ and $\mathbf{S} \in \{0, 1\}^{Z \times Z}$ represent the input patient node feature matrix and the patient adjacency matrix of the patient weight network. Formally, $\mathbf{W}^l$ and $\{h_a^{(l+1)}\}_{a=1,...Z}$ represent the weight matrix and features (embeddings) in the $l$th layer of node a in the CGRL model.

Our purpose is to predict whether the patient is at risk of developing the target disease. $L_{cl}$ represents the loss function of disease classification, $\hat{y}$ represents the predicted probability, and $p_i$ is the representation of the patient.

## 4. Methodology

In this section, we elaborate on the method of constructing the patient network and the differential comorbidity network based on the comorbidity network, which are the cornerstones for hypertension comorbidity risk prediction. Then, how to construct the proposed Risk prediction learning method for hypertension comorbidity is introduced. We finally constructed the CGRL method. Fig. 1. illustrate the CGRL framework.

*4.1 Constructing the patient network based on common diseases*

This section introduces the method of constructing a patient weight network using edge weights. The edge weights represent the number of common diseases between different patients. Given the complexity of patient correlation, constructing a patient-weighted network by selecting the appropriate threshold is needed.

The method of constructing patient networks and evaluating similarity involves calculating the frequency of common diseases. The number of diseases shared by two patients was used to measure patient similarity. The elements of the matrix are the recorded number of patients and the number of diseases. The idea is to use graph theory concepts and metrics to construct a bipartite graph to represent the relationship between patients and diseases. If the weights between patients are the number of neighbors with the same disease, it is called "weighted patient network".

*4.2 Constructing the co-occurrence correlation network based on the comorbidity network*

To better explore the comorbidity community structure, we define the comorbidity relationship between two diseases as the comorbidity frequency of the common diseases. The constructed comorbidity network is an undirected graph in which a disease represents a node. The co-occurrence correlation (COCO) method is widely used to measure the correlation between comorbidities, as it does not have many limitations. We use it to quantify the correlation between two diseases [4], [8], [9]. It was calculated as follows:

$$\text{COCO}_{ixjx} = \frac{\beta \text{CO}_{ixjx}}{\sqrt{\text{PR}_{ix}^2 + \text{PR}_{jx}^2}}. \tag{1}$$

wherein $CO_{ixjx}$ is the co-occurrence of $ix$ and $jx$, $\beta$ usually takes a value of $\sqrt{2}$, and $PR_{ix}$ and $PR_{jx}$ represent the prevalence of diseases $ix$ and $jx$.

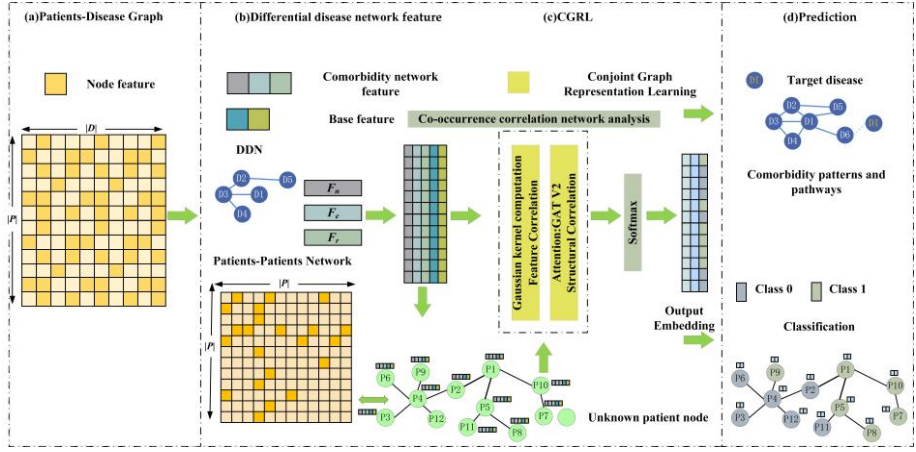

Fig. 1. Illustration of the proposed CGRL framework.

*4.3 Network feature extraction method based on differential comorbidity network*

Traditional machine learning for patient risk disease prediction only considers the embedding of disease features, without considering the comorbidity network features as input features, making it difficult to better explain the relationship between comorbidity networks and patient risk disease prediction. In this study, to capture the relationship between patients and diseases in a more refined way, we proposed a differential comorbidity network technique to explore the potential disease risk of patients.

To solve the problems mentioned above, a differential network has been constructed, and features have been extracted based on prior knowledge. The network, which was constructed using training set data, integrated a large network of disease information. This large network was constructed
by using 65% of the samples as prior knowledge. Each patient had their small network constructed using only their information. The network features were calculated based on the similarity between their small network and the large network. The remaining 35% of the samples were used by the CGRL method.

With the collected data mentioned previously, we are now able to construct the differential network that can characterize hypertension. Inspired by Hossain [56] and Hang Qiu *et al.* [4], [5], [9], three methods for calculating differential network features were proposed, including a network feature node scoring method based on disease vector similarity. As a feature based on disease similarity, node scoring measures the relationship between an individual patient's disease network (PN) and a differential disease network (DDN) from
the perspective of disease similarity. The calculation method of node features is based on the cosine similarity transformation. The disease node score is defined as the following equation:

$$F_n = \frac{\sum_{i=1}^{N}(d_{iPN}) * f(d_{jDDN})}{\sum_{i=1}^{Na}(d_{iPN})}. \quad (2)$$

wherein $d_i$ is the network node and $N$ is the total number of common nodes or diseases between DDN and PN.

The edge score feature denotes another feature based on disease vector similarity. In the constructed differential comorbidity network, edges represent the co-occurrence relationships between diseases. By calculating the similarity of the edge vectors of PN and DDN, edge scores are obtained to describe the differences. The edge score of a patient (i.e., PN) is defined as follows:

$$F_e = \frac{\sum_{i=1, j=1, i \neq j}^{N} e_{(d_i,d_j)PN} * f(e_{(d_i,d_j)DDN})}{\sum_{i=1, j=1, i \neq j}^{Te} e_{(d_i,d_j)PN}}. \quad (3)$$

wherein $e$ is the edge of the network and $Te$ is the total number of edges of the PN.

By applying the PageRank algorithm, the node ranking score of a patient is defined as follows:

$$F_r = \frac{\sum_{i=1}^{N}(d_{iPN}) * pk(d_{jDDN})}{|D(PN)|}. \quad (4)$$

wherein $pk(d)$ denotes the PageRank value. $|D(PN)|$ denotes the total number of disease nodes in PN.

*4.4 Risk prediction learning method for hypertension comorbidities*

Exploring the relationship between hypertension comorbidities and complications is crucial for understanding their interaction mechanisms, screening, and synergistic treatment. However, due to the complexity of comorbidities, this goal still has a long way to go. Although various medical data and methods have been used to study comorbidities, the exploration of comorbidities through graph learning methods combined with comorbidity, patient risk disease prediction, and comorbidity network analysis is still limited. Conventional machine learning methods face the dilemma of high dimensionality. To discover potential relationships with the GNN method, some scholars have proposed many methods [56]-[60] to tackle this problem and have obtained some valuable results. However, it does not consider the attention of features, which can lead to poor classification performance and urgently needs more accurate and effective prediction models.

To capture the dependencies among the embedded representations of individual patients and features (basic and comorbidity differential features), we proposed a CGRL method in which the attention coefficient is calculated by combining features and structures [61]-[71]. The similarity of features is calculated by Gaussian kernel approximation. The adaptive attention coefficient is added in the GATV2 structure, the input dimension is the comorbidity network feature, and the output dimension is the embedding dimension of the new patient node exported by the model after learning and passing through the neural network. In the prediction of comorbidities in hypertension, the patient disease bipartite graph is used to aggregate the information transmitted by patient nodes and disease nodes through a joint information transmission mechanism, update the node embedding vector representation, and thus more effectively improve the probability of predicting comorbidities in hypertension patients. The problem can be expressed as follows [61]:

$$\mathbf{C}_{ab} = \arg\min_{\mathbf{VV}_{ab}^T} \left( \mathbf{S}_{ab} - \sum_b \mathbf{VV}_{ab}^T \mathbf{S}_{ab} \right)^2. \tag{5}$$

wherein $\mathbf{C}_{ab}$ is a learned structural intervention parameter that represents the degree of structural intervention between node $a$ and node $b$. The contextual correlation between two connected nodes, which are obtained through Gaussian kernel computation:

$$g_{ab} = \frac{1}{\sqrt{2\pi}\sigma} \exp\left( -\frac{1}{2\sigma^2} \left\| \mathbf{W}^l \mathbf{h}_a^l - \mathbf{W}^l \mathbf{h}_b^l \right\|^2 \right), \tag{6}$$

In equation (6), the conjoint message-passing layer processes a set of node features $\{h_a^l\}_{a=1,...Z}$. This layer maps these features into a $D^{l+1}$ dimensional space $\{h_a^{(l+1)}\}_{a=1,...Z}$. $\mathbf{W}^l$ is the weight matrix of the feature mapping. $C_{ab}$ is the learned structural intervention parameter, representing the degree of structural intervention between the patients' node $a$ and node $b$. The calculation of $t_{ab}$ is as follows:

$$t_{ab} = \frac{\exp(\mathbf{C}_{ab})}{\sum_{k \in Z_a} \exp(\mathbf{C}_{ak})}. \tag{7}$$

Each layer introduces two learnable parameters $r_o$ and $r_t$ to determine the relative significance between the patients' features and structural correlations:

$$r_o = \frac{\exp(g_o)}{\exp(g_t) + \exp(r_o)}, r_t = \frac{\exp(g_t)}{\exp(g_t) + \exp(g_o)}, \tag{8}$$

Where $r_o$ or $r_t$ represents the normalized significance related to different types of correlations. Compute the attention score $\delta_{ab}$:

$$\delta_{ab} = \frac{r_o \cdot o_{ab} + r_t \cdot t_{ab}}{\sum_{k \in Z_a}[r_o \cdot o_{ak} + r_u \cdot t_{ak}]} = r_o \cdot o_{ab} + r_t \cdot t_{ab}. \tag{9}$$

Eq. (5) describes a method for computing the patients' attention mechanism between the patients' nodes in graph neural networks (GNNs), based on both the patients' structural and node-to-node correlation features. After obtaining the joint attention scores, the linear combination of features corresponding to each node and its neighbors is computed as the output. In graph neural networks (GNNs), where $\epsilon \in (0,1)$ is a learnable parameter that improves the expressive capability. The calculation of the feature representation $h_a^{l+1}$ of node $a$ in the next layer $l+1$ is as follows:

$$\mathbf{h}_a^{l+1} = \left( \delta_{aa} + \varepsilon \cdot \frac{1}{|Z_a|} \right) \mathbf{W}^l \mathbf{h}_a^l + \sum_{b \in Z_a, b \neq a} \delta_{ab} \mathbf{W}^l \mathbf{h}_b^l. \tag{10}$$

The final representation of the patient is output from CGRL through the linear layer and the Softmax activation function to output the disease classification and predict the outcome. The calculation formula is as follows:

$$\hat{y} = softmax(W_y p_i + b_y), \tag{11}$$

To achieve the goal of disease classification, the classification loss of risk diseases is calculated based on cross-entropy. The loss function is as follows:

$$L_{cl} = -\frac{1}{Z} \sum_{i=1}^{Z} \left( y_i \log(\hat{y}_i) + (1 - y_i) \log(1 - \hat{y}_i) \right). \tag{12}$$

Wherein $L_{cl}$ represents the loss function of disease classification, $\hat{y}$ represents the predicted probability, and $p_i$ is the representation of the patient.

Table 1. Dataset details of the cardiovascular metabolic disease.

| Dataset | Nodes | CL | Description |
|---|---|---|---|
| DM | 1024 | 2 | Hypertension with DM |
| CHD | 1668 | 2 | Hypertension with CHD |

Table 2. Comparison of classification effects of different prediction methods by 5 rounds.

| Method | DM Acc | DM F1 | CHD Acc | CHD F1 |
|---|---|---|---|---|
| APPNP | $0.7000_{\pm 0.0738}$ | $0.6982_{\pm 0.0737}$ | $0.7237_{\pm 0.0249}$ | $0.7215_{\pm 0.0264}$ |
| DGI | $0.7444_{\pm 0.0076}$ | $0.7432_{\pm 0.0069}$ | $0.7373_{\pm 0.0158}$ | $0.7356_{\pm 0.0170}$ |
| GSAGE | $0.7194_{\pm 0.0181}$ | $0.7163_{\pm 0.0172}$ | $0.7068_{\pm 0.0221}$ | $0.7038_{\pm 0.0255}$ |
| SGC | $0.7000_{\pm 0.0286}$ | $0.6949_{\pm 0.0313}$ | $0.7390_{\pm 0.0395}$ | $0.7386_{\pm 0.0393}$ |
| GCN | $0.7472_{\pm 0.0181}$ | $0.7456_{\pm 0.0178}$ | $0.7169_{\pm 0.0195}$ | $0.7125_{\pm 0.0205}$ |
| JKNET | $0.7278_{\pm 0.0335}$ | $0.6863_{\pm 0.0319}$ | $0.7322_{\pm 0.0129}$ | $0.7274_{\pm 0.0128}$ |
| GAT | $0.7028_{\pm 0.0158}$ | $0.6827_{\pm 0.0257}$ | $0.7017_{\pm 0.0211}$ | $0.6773_{\pm 0.0103}$ |
| **CGRL** | $\mathbf{0.7945_{\pm 0.0181}}$ | $\mathbf{0.7894_{\pm 0.0212}}$ | $\mathbf{0.7593_{\pm 0.0047}}$ | $\mathbf{0.7586_{\pm 0.0041}}$ |

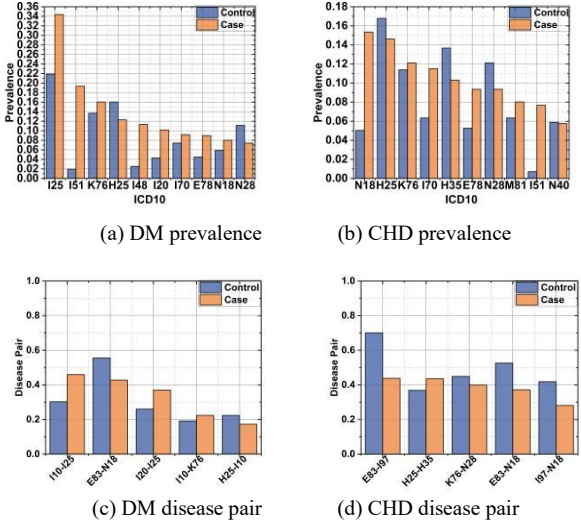

(a) DM prevalence  (b) CHD prevalence

(c) DM disease pair  (d) CHD disease pair

Fig. 2. Comparison of the ratio between disease prevalence and disease node pairs.

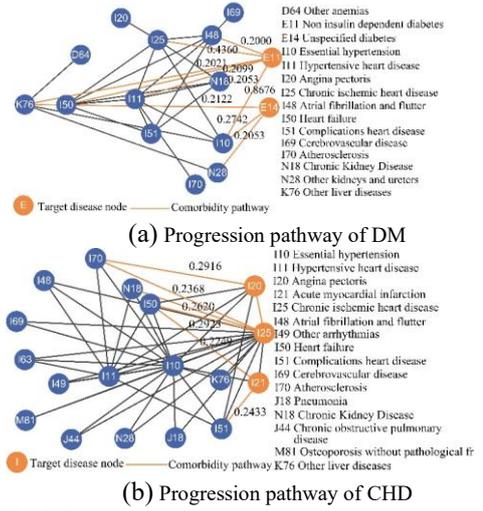

(a) Progression pathway of DM

(b) Progression pathway of CHD

Fig. 3. Target disease progression pathway.

## 5 Results

In this section, we introduce the setup of experiments and validate the effectiveness of CGRL in hypertension comorbidity risk prediction. Moreover, to discover the comorbidity patterns between different diseases, we also conduct a comorbidity network analysis experiment.

*5.1 Datasets description and processing*

This study has been approved by the hospital ethics committee, and the datasets have been anonymized and processed. The research is a retrospective case study, and informed consent from patients was exempted. The dataset is from the medical record homepage, includes admission and discharge time, disease diagnosis, etc., and the time range is from 2020 to 2023 by discharge time.

To predict the risk of disease, two groups of patients have been selected in Table 1. Inclusion and exclusion criteria are as follows: first, patients diagnosed with DM and CHD for the first time and then hypertension or diabetes and hypertension at the same time were excluded; second, patients with less than two hospitalization records were further excluded. Among them, patients who were diagnosed with hypertension first and then diagnosed with DM and CHD were taken as the case group, while other patients who were not diagnosed with DM and CHD were taken as the control group.

*5.2 Experimental settings and evaluation metrics*

The experiments have been conducted on a PC with a 2.5 GHz i9 CPU and 32GB RAM. All the methods have been implemented in Python 3.8 to test whether they are suitable for medical datasets. The datasets were divided into the training set, validation set, and test set of 0.6:0.2:0.2 ratio.

Note that the hyperparameters of $\lambda$, $\eta$, $hi$, $dr$, and $hd$ denote step size, learning rate, hidden layer dimension, dropout and heads, respectively. In the DM and CHD datasets, $\lambda$, $\eta$, $dr$, and $hd$ are 0.01, 0.02, 0.06, and 8, respectively. The hidden layer dimension $hi$ is 128 in the DM dataset and 8 in the CHD dataset. On all datasets, we run each baseline five times to obtain the average performance. To evaluate the classification performance of all approaches, we adopt Accuracy (ACC) and F1-score (F1) as evaluation metrics.

*5.3 Comparison baselines*

To illustrate the effectiveness of CGRL, the comparison experiments have been completed. We compare the performance of the CGRL model with the strong GNN baselines: APPNP, DGI, GraphSAGE, SGC, GCN, JKNET, and GAT. We use official codes for all baseline GNN models and configure parameters using the official recommended settings. Existing baseline is not limited to existing prognostic prediction models, and all of these strong GNN models lack attention to information for structural interventions of patient interactions. All these comparative models did not consider obtaining features from disease nodes and edges, and the importance of ranking disease nodes in the comorbidity network. They failed to establish a direct connection between the patient and comorbidity networks, lacking interpretability.

*5.4 Classification performance comparisons*

In this part of the experiment, we evaluate CGRL and compare its performance with strong GNN baselines. The average classification performance evaluated by ACC and F1-score has been listed in Table 2. When evaluated by ACC, CGRL outperforms the best baseline (GCN) by 6.33% and outperforms DGI by 6.73% on the DM dataset. In the CHD dataset, CGRL outperforms the best baseline (SGC) by 2.75% and outperforms DGI by 2.98%, respectively. When evaluated by F1, CGRL outperforms the best baseline (GCN) by 5.87% and outperforms DGI by 6.21% on the DM dataset.
In the CHD dataset, CGRL outperforms the best baseline (SGC) by 2.71% and outperforms DGI by 3.13%, respectively.

To further discover the comorbidity patterns between different diseases, in an attempt to understand the developmental pathways and the underlying pathological mechanisms between diseases more clearly, we conducted a comorbidity network analysis experiment.

*5.5 Findings by analysis of the comorbidity networks*

We found the risk for DM patients associated with the high-risk clusters was larger in the case group than in the control group. The pathological mechanism of diabetes is a metabolic disorder. Clarifying the cause of the disease helps explain the pathogenesis. It is found that the prevalence of chronic ischemic heart disease (I25, 34.38%), complications of heart disease (I51, 19.34%), and other diseases of the liver (K76, 16.02%) in the case group are higher than the control group in Fig.2 (a), suggesting that these disease nodes may play an important role in the development of diabetes. We also found that the prevalence of chronic kidney disease (N18, 15.35%), senile cataract (H25, 14.63%) in the case group is higher than in the control group in Fig.2 (b), suggesting that these disease nodes may lead to coronary heart disease. Chronic ischemic heart disease (I25) and primary hypertension (I10) have a common disease relationship in Fig.2 (c), indicating that coronary heart disease is also related to the occurrence of diabetes, which is consistent with the current report [19]-[21]. We identified the high-risk clusters through which many patients develop into the CHD. The ratio of disease pairs in the CHD group is different. Fig. 2 (d) shows that the ratio of interaction of senile cataracts and other retinal diseases (H25-H35, 0.44), the case group is stronger than the control group. Among them, senile cataracts and essential hypertension (H25-I10, 0.23) are lower in the combined case group than in the control group. As part of the western region of China, we compared our research results with those of Sichuan Province and confirmed our previous findings through the comparison.

To show the critical path, we selected paths with a probability threshold greater than 0.2 to highlight (orange line). For instance, as shown in Fig. 3, the numerical values represent the comorbidity probability. The pathway to the confirmed target disease in their comorbidity network is determined based on the diagnosis of high-risk patients. As shown in Fig. 3 (b), evidence shows that atherosclerosis (I70), hypertensive heart disease (I11), and chronic kidney disease (N18) may induce coronary heart disease. The previously reported evidence shows that different coronary heart diseases (I20, I25, and I21) interact with each other [72].

# 6 Conclusion

In this paper, we have proposed the Conjoint Graph Representation Learning (CGRL) framework for detecting diabetes and cardiovascular comorbidity based on the hypertension comorbidity network feature. Compared with strong GNN models, CGRL achieves high prediction accuracy in the real medical dataset. Moreover, it can discover and quantify latent correlations among diseases, and identify the differential disease nodes and node pairs in the pathway of DM and CHD, thus providing an early warning of comorbidities.

In the future, we will design a new model with a dynamic perception structure and features to improve the predictive performance of comorbidity progression as follows: first, other similarity measures can also be attached to construct the patient network and the comorbidity network. To establish a patient network based on similarity methods, such as the Jaccard index, not only based on the weight model of the patient network. Second, in addition to the coefficients learned in the joint attention computation, we will also explore more coefficients calculated in heterogeneous patient graphs and time series graph data. Since the input vector is scalable, we also collect the clinical test features of patients and take the similarity of disease test results as the target of the model.